\documentclass[letterpaper]{article} 
\usepackage{aaai23}  
\usepackage{times}  
\usepackage{helvet}  
\usepackage{courier}  
\usepackage[hyphens]{url}  
\usepackage{graphicx} 
\usepackage{natbib}  
\usepackage{caption} 
\usepackage{algorithm}
\usepackage{algorithmic}

\usepackage{newfloat}
\usepackage{listings}
\DeclareCaptionStyle{ruled}{labelfont=normalfont,labelsep=colon,strut=off} 
\floatstyle{ruled}
\newfloat{listing}{tb}{lst}{}
\floatname{listing}{Listing}
\usepackage{graphicx}
\usepackage{amsmath}
\usepackage{amssymb}
\usepackage{booktabs}

\usepackage{booktabs}
\usepackage{tabularx}

\usepackage{multirow}
\usepackage{diagbox}

\usepackage{cuted}
\usepackage{capt-of}
\usepackage{pifont}

\title{Spatiotemporal Deformation Perception for Fisheye Video Rectification}
\author{
    Shangrong Yang,
    Chunyu Lin\thanks{Corresponding author: cylin@bjtu.edu.cn},
    Kang Liao,
    Yao Zhao
}

\affiliations{

    Institute of Information Science, Beijing Jiaotong University\\
    Beijing Key Laboratory of Advanced Information Science and Network, Beijing, 100044, China\\
    \{sr\_yang, cylin, kang\_liao, yzhao\}@bjtu.edu.cn 
}

\usepackage{bibentry}

\begin{document}

\maketitle

\begin{abstract}
Although the distortion correction of fisheye images has been extensively studied, the correction of fisheye videos is still an elusive challenge. For different frames of the fisheye video, the existing image correction methods ignore the correlation of sequences, resulting in temporal jitter in the corrected video. To solve this problem, we propose a temporal weighting scheme to get a plausible global optical flow, which mitigates the jitter effect by progressively reducing the weight of frames. Subsequently, we observe that the inter-frame optical flow of the video is facilitated to perceive the local spatial deformation of the fisheye video. Therefore, we derive the spatial deformation through the flows of fisheye and distorted-free videos, thereby enhancing the local accuracy of the predicted result. However, the independent correction for each frame disrupts the temporal correlation. Due to the property of fisheye video, a distorted moving object may be able to find its distorted-free pattern at another moment. To this end, a temporal deformation aggregator is designed to reconstruct the deformation correlation between frames and provide a reliable global feature. Our method achieves an end-to-end correction and demonstrates superiority in correction quality and stability compared with the SOTA correction methods.
\end{abstract}

\section{Introduction}

Many existing algorithms including object detection \cite{Redmon2016YouOL}\cite{Lin2017FeaturePN} and tracking \cite{tracking}\cite{Li2019TargetAwareDT}, pedestrian re-identification \cite{Pedagadi2013LocalFD}, and human pose estimation \cite{Newell2016StackedHN} are designed based on the perspective camera models. Although they achieve satisfactory performance on ordinary planar video, their field-of-view (FOV) are limited. As a result, in most scenarios, fisheye cameras replace multiple perspective cameras to obtain a wider field-of-view (FOV) at a cheap cost. However, the structure of the fisheye video is different from that of the distorted-free video. If these algorithms are directly applied to fisheye video, the performance will be dramatically degraded. Therefore, it is necessary to pre-correct the video to maintain the performance of subsequent algorithms.

\begin{figure}[!t]
    \centering
    \includegraphics[scale=0.112]{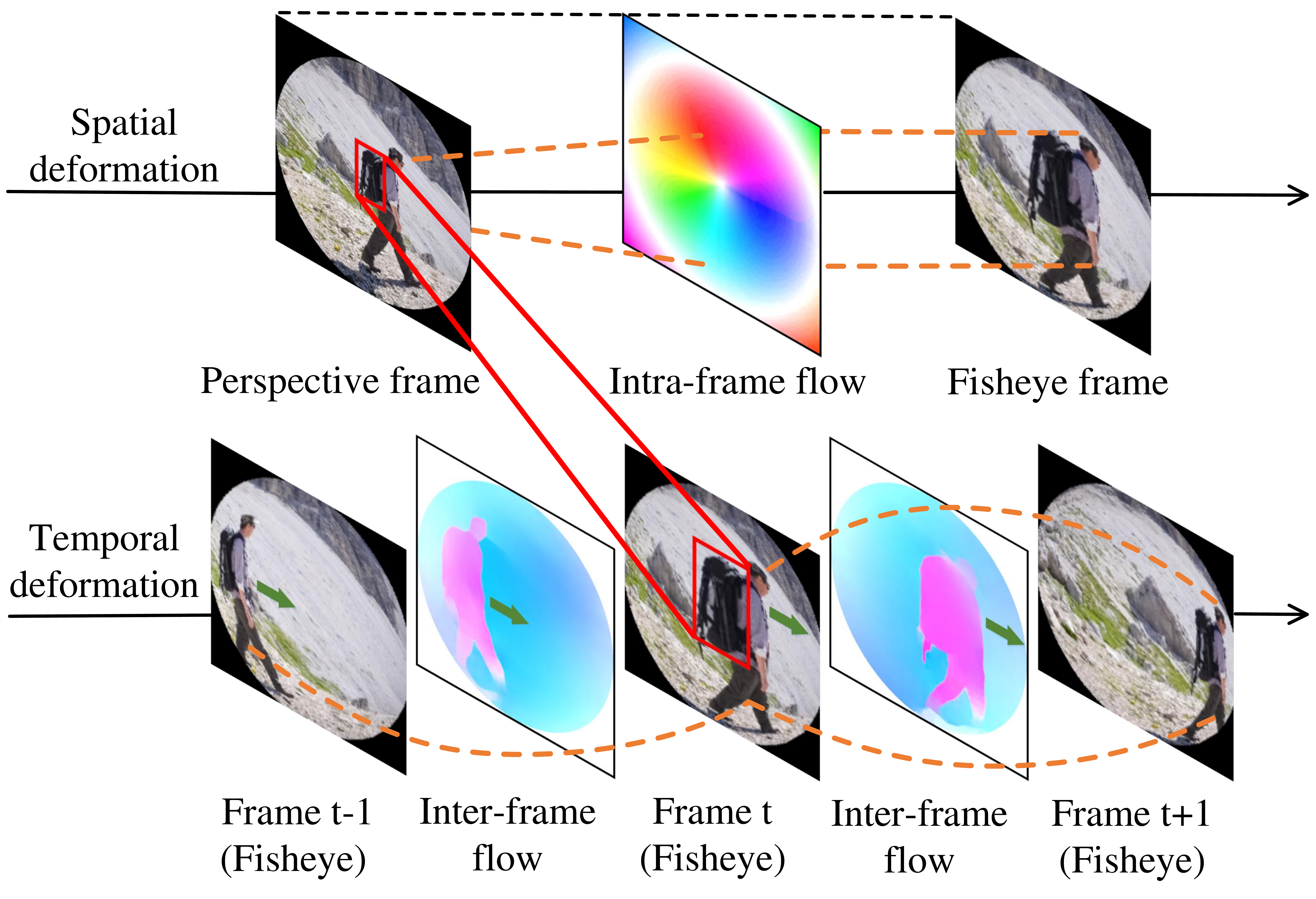}
    \caption{
    \label{intro_img}
    The spatial and temporal deformation. The orange line represents the pixel motion caused by the distortion. The red box indicates that it is possible for the distorted moving object to find its distorted-free appearance at other times.}
\vspace{-0.3cm}
\end{figure}

The correction methods for video processing have rarely been investigated, and most of them are based on the image processing. A series of traditional methods \cite{Reimer2013INTCAL1A}\cite{Reimer2009Intcal09AM}\cite{ZHANG1999} correct fisheye images with the help of calibration boards. They require labor-intensive manual operation. Many self-calibration methods \cite{Dansereau2013DecodingCA}\cite{Rui2014Unsupervised} can automatically detect significant features to calculate the distortion parameters. However, the detected features are greatly affected by the scene content. With the development of deep learning, many novel networks have been designed for distortion correction. \cite{Rong2016Radial}\cite{DeepCalib}\cite{Yin2018FishEyeRecNet}\cite{Xue2019} improve the convolutional neural networks (CNNs) to regress the distortion parameters. \cite{Liao2019}\cite{DDM}\cite{Blind} design the network on the architecture of generative adversarial networks (GANs) to generate rectified images directly. These methods are specially devised for image rectification and do not consider the temporal correlation of fisheye video frames. Different frames generate different results, which cause instability in the corrected video. Achieving consistent correction for each frame is the biggest challenge for existing correction methods.

We regard the pixel motion caused by the distortion of a single image as \textit{intra-frame} optical flow and the motion between different frames as \textit{inter-frame} optical flow. In this paper, we propose a fisheye video correction method based on spatiotemporal deformation perception\footnote{Available at https://github.com/uof1745-cmd/SDP}. Considering the output of a stable algorithm should not change significantly due to the input of new frames, we first predict the intra-frame optical flow independently for each frame. Then we use a temporally decreasing weighting scheme to generate a coarse intra-frame optical flow to mitigate the impact of new input. However, independent prediction for intra-frame optical flow ignores the temporal correlation. Since the distortion of the fisheye video increases with the radius, it is possible for the distorted moving object to find its distorted-free appearance at other times, as shown in Figure \ref{intro_img}. Therefore, we introduce a temporal deformation aggregator to reconstruct the temporal relationship between the video frames. The reconstructed global feature is conducive to predicting a more accurate intra-frame optical flow. In addition, according to the derivation, we found dual optical flows (the inter-frame optical flow of fisheye video and its corresponding flow of distorted-free video) can effectively reflect object's spatial deformation. We thereby use the independent intra-frame optical flow to correct the corresponding fisheye frame and estimate dual optical flows. A spatial local deformation attention is then calculated from dual optical flows to enhance the local accuracy of the intra-frame optical flow. 

In summary, this work includes the following contributions:

\begin{itemize}
    \item We pioneer to explore the stable correction on fisheye video and propose a temporal weighting scheme to alleviate the jitter problem in existing correction methods.

    \item Our network fully perceives the spatiotemporal deformation of objects, which is leveraged to simultaneously enhance the global and local accuracy of the prediction.

    \item Compared with existing methods, our method fully exploits the spatiotemporal information of the video and achieves promising improvements in both video stability and correction quality.
\end{itemize}

\section{Related Work}
Distortion correction is a meaningful work because it greatly alleviate the impact of distortion on algorithms that are designed for planar cameras, such as object detection \cite{Ren2015FasterRT}\cite{Lin2020FocalLF}, semantic segmentation \cite{Girshick2014RichFH}\cite{Shelhamer2017FullyCN} and pedestrian re-identification \cite{Liu2015ASA}\cite{Han2021KISSFR}. Most of the existing distortion correction methods \cite{Mei2007}\cite{Rui2014Unsupervised}\cite{Bukhari2013Automatic} are designed based for images. At the beginning, distortion correction was simplified to the camera calibration \cite{Zhang2000AFN}\cite{Luong2004SelfCalibrationOA}. However, the method fails on fisheye images for two reasons: (1) Ordinary pinhole camera models cannot express the relationship between fisheye images and 3D scenes; (2)Calculating the parameters of the fisheye camera model \cite{sphere_model}\cite{Basu1995}\cite{Aleman2014Automatic} requires more detected features. Therefore, Mei et al. \cite{Mei2007} proposed a manual calibration method. He leveraged line image points to calculate the distortion parameters. Melo et al. \cite{Rui2014Unsupervised} designed an automatic method and greatly improved the smoothness of distortion correction. Nevertheless, in complex scenes, the feature selection has low accuracy, which further affects the performance of correction.

Considering that the neural network can automatically locate the most representative features, many researchers used neural networks to extract features. Rong et al. \cite{Rong2016Radial}, Yin et al. \cite{Yin2018FishEyeRecNet}, Xue et al. \cite{Xue2019} leveraged convolutional neural networks(CNNs) to predict parameters. Although their performance exceeded self-calibration methods \cite{Rui2014Unsupervised}, it was difficult to accurately predict all the parameters. Another researcher used Generative Adversarial Networks (GANs) for distortion correction. They reasoned that the domain gap between the distorted and corrected images is smaller than that between the image and the parameters. Liao et al. \cite{Liao2019} \cite{DDM}, Yang et al. \cite{PCN}, Zhao et al. \cite{Zhao2021RevisitingRD} used GANs to implement the transformation from fisheye image to normal image. Compared with the regression method, their correction speed has been largely improved, but the generated content has artifacts.

The aforementioned methods are designed for image correction, while the distortion video correction method is under-explored. Since the video has one more temporal dimension than the image. Therefore, using the image approaches to correct video frame by frame has two problems: (1) Correcting each frame individually, the network produces different outputs due to the varying input content and generates a corrected video with jitter effect. (2) Rectifying the video with uniform transformation, it is difficult to find the best predicted parameter. Therefore, in this paper, we take the temporal dimension into account and design a deep video correction method. Our method solves the jitter problem as well as improves the accuracy of the correction.

\section{Fisheye Model}
The camera model reflects the mapping process from a 3D scene to an image. There are three models commonly used to express the mapping relationship of fisheye cameras: the sphere model \cite{sphere_model}, the polynomial model \cite{Basu1995} and the division model \cite{Aleman2014Automatic}. The sphere model indicates an excellent fit between the panoramic scene and the fisheye image. However, compared with panoramic images, perspective images are easily accessible. Therefore, many researchers leveraged the division model or the polynomial model to achieve a simpler transformation. Since the polynomial model is similar to the division model and has been frequently used in recent years. Therefore, we leverage the polynomial model to construct our dataset. Generally, the polynomial model can be expressed as
\begin{equation}
  \begin{bmatrix}x_{u}\\ y_{u}\end{bmatrix}=(1+k_{1}{r_{d}}^{2}+k_{2}{r_{d}}^{4}+k_{3}{r_{d}}^{6}+\cdots )\begin{bmatrix}x_{d}\\ y_{d}\end{bmatrix}
\end{equation}

Where $(x_{u},y_{u})$ and $(x_{d},y_{d})$ are the image coordinates on the perspective image and fisheye image. $\left [ k_{1},k_{2},k_{3},\cdots \right ]$ are distortion parameters, which can control the distortion degree of the image. The distortion radius is represented by $r_d$, which is the distance between the distortion coordinates $(x_d, y_d)$ and the distortion center $(x_0, y_0)$. It can be calculated as

\begin{equation}
  r_{d}= \sqrt{\left ( x_{d}-x_{0} \right )^{2}+\left ( y_{d}-y_{0} \right )^{2}}
\end{equation}

\begin{figure}[!t]
  \centering
  \includegraphics[scale=0.16]{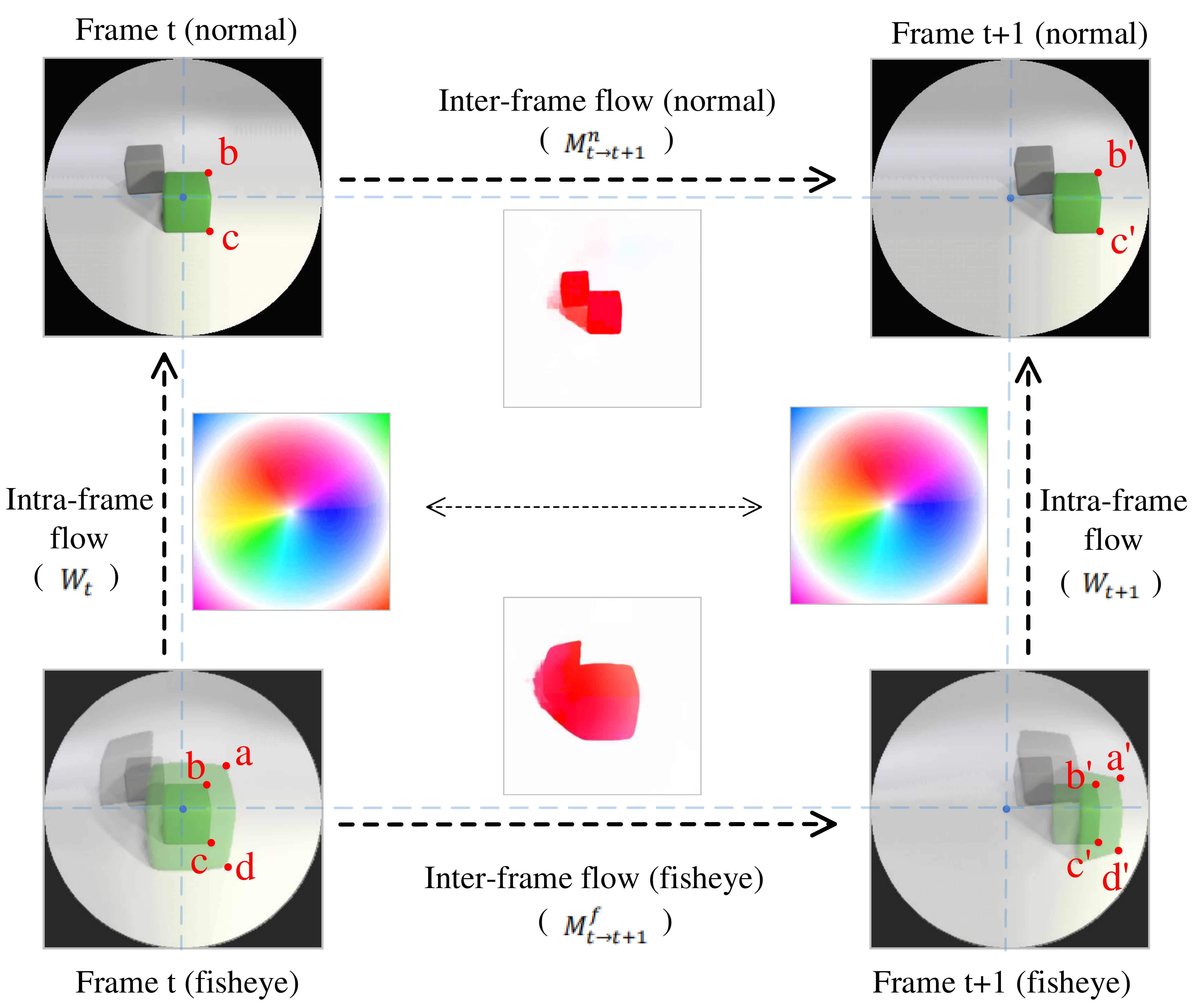}
  \caption{
  \label{inter_intra_flow}
  The relationship between the inter-frame flow and the intra-frame flow. The blue dotted lines indicate the alignment of the image center.}
\vspace{-0.3cm}
\end{figure}

\section{Dual Optical Flow and Fisheye Deformation}
\label{Dual Optical Flow}
For video sequences, if an object is located in the center of the image in the first frame, it gradually moves towards the boundary in subsequent frames. Since the distortion of fisheye images increases with radius, we can observe that the object gradually deforms. However, even though the object has some displacement, it will have no deformation in the perspective video. By perspective video, we want to denote the video is captured with a perspective camera with no obvious deformation in contrast with fisheye lens. We consider using the displacements of objects in fisheye video and distorted-free video to infer the deformation. As shown in Figure \ref{inter_intra_flow}, suppose there is a square object at frame $t$ \cite{Clevr_img} of a perspective video, and its upper right corner and lower right corner are located at pixel points $b$ and $c$, respectively. If the object has translation, the upper right and lower right corners will move to ${b}'$ and ${c}'$ at frame $t+1$. Because the object deforms in fisheye video, distortion points $a$ and $d$ correspond to $b$ and $c$, respectively. The distortion points of ${a}'$ and ${d}'$ correspond to ${b}'$ and ${c}'$ at frame $t+1$. We use $M^{f}$ and $M^{n}$ to denote the optical flow map of fisheye video and distorted-free video, respectively. $M_{t \to t+1 }^{f}(x)$ represents the value of $M^{f}$ at the pixel point $x$ in frame $t$. $M_{t \to t+1 }^{n}(x)$ is the pixel point value of $M^{n}$ in frame $t$. $L(x)$ is the pixel coordinate of the $x$, then we have
\begin{equation}
  M_{t \to t+1 }^{f}(a)=\overrightarrow{{a}'a}=L({a}')-L({a})
\end{equation}
\begin{equation}
  M_{t \to t+1 }^{n}(b)=\overrightarrow{{b}'b}=L({b}')-L({b})
\end{equation}

By subtracting these two formulas, we can obtain:
\begin{equation}
  \label{formalu_flow}
M_{t \to t+1 }^{f}(a)-M_{t \to t+1 }^{n}(b)=(L({a}')-L({b}'))-(L({a})-L({b}))
\end{equation}

We denote the intra-frame optical flow which is the pixel motion caused by the distortion of fisheye video to $W$. It represents the uniform deformation performed on each fisheye frame. The value of $W$ at the point $x$ in the $t$ frame is $W_{t}(x)$. $W_{t}({a}')$ and $W_{t}(a)$ can be specifically expressed as
\begin{equation}
  W_{t}({a}')=L({a}')-L({b}')
\end{equation}
\begin{equation}
  W_{t}(a)=L(a)-L(b)
\end{equation}

Then, Equation \ref{formalu_flow} can be simplified to
\begin{equation}
\label{deta}
\Delta M= M_{t \to t+1 }^{f}(a)-M_{t \to t+1 }^{n}(b)=W_{t}({a}')-W_{t}(a)=\Delta W
\end{equation}

The difference in the intra-frame optical flow can be obtained by subtracting the flow map of the distorted-free video from the corresponding fisheye flow map. The difference between the fisheye and the normal flow map represents the deformation of the object during motion. Therefore, the flow map of fisheye video and preliminary corrected video can benefit the network to perceive the deformation of the object, thereby enhancing the local accuracy of intra-frame optical flow.

\begin{figure*}[!t]
\centering
\includegraphics[scale=0.113]{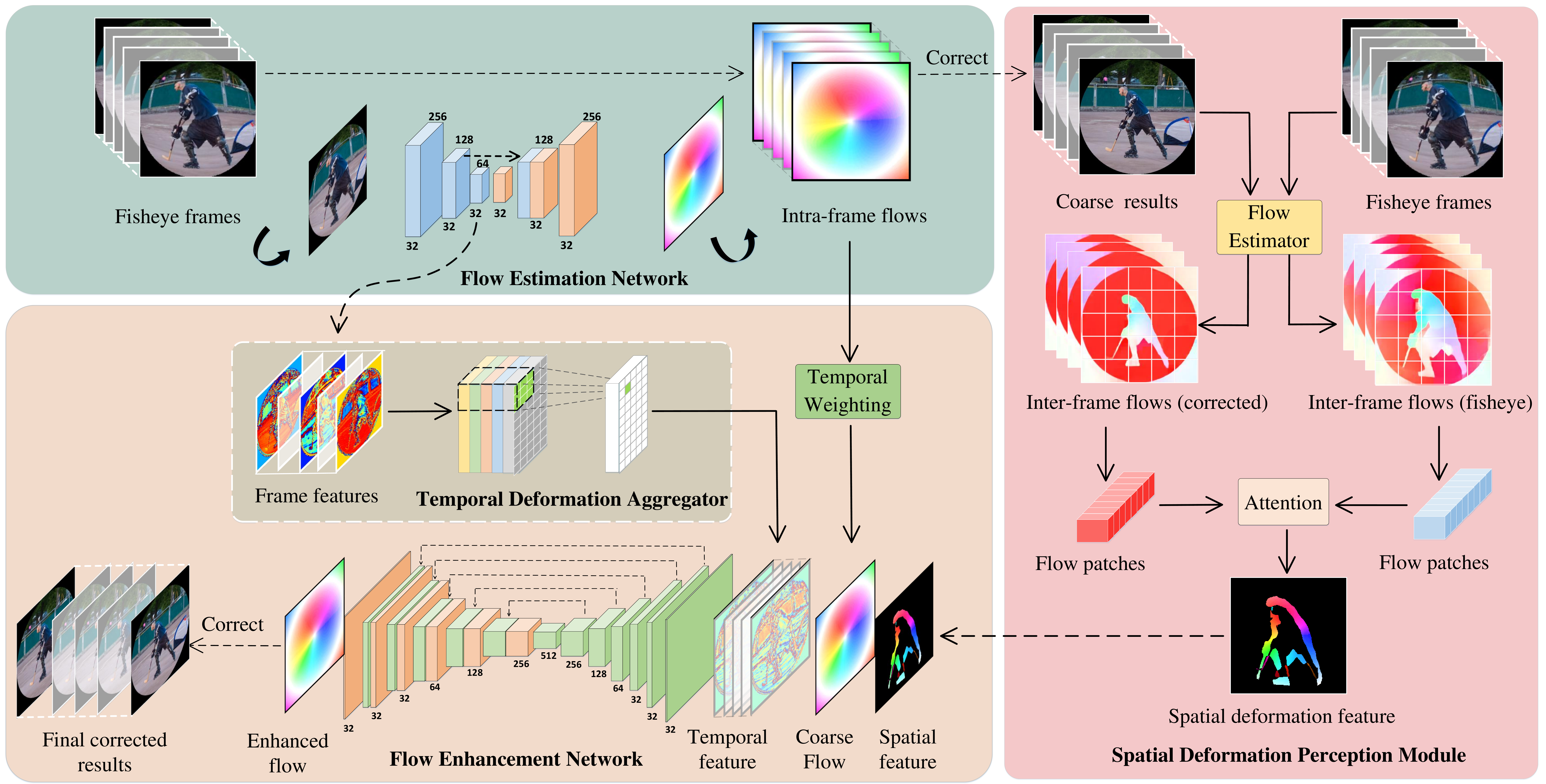}
\caption{
\label{structure}
The overview of our network. The flow estimation network predicts intra-frame flows for preliminary correction. A temporal weighting scheme is first used to alleviate the video jitter. Then the extracted features and the preliminary correction results are sent to the temporal deformation aggregator and spatial deformation perception module to establish spatiotemporal correlation. The flow enhancement network benefits from all useful information and generates accurate results.}
\vspace{-0.3cm}
\end{figure*}

\section{Architecture}
Compared with image correction, video correction is more challenging because of the additional temporal dimension. To this end, we propose a multi-stage correction network to maintain the stability of corrections in timestamps and achieve smooth transitions between different timestamps. Our multi-stage correction network includes the flow estimation network (FSN) and the flow enhancement network (FEN), as shown in Figure \ref{structure}. The FSN performs a simple estimation of the intra-frame optical flow for each fisheye frame. The FEN removes the jitter effect with a temporal weighting scheme. In addition, a temporal deformation aggregator (TDA) is presented to enhance global correlation, and a spatial deformation perception module (SDM) is used to enhance local accuracy. Finally, the global optical flow generated by the FEN is used to resample the fisheye video to obtain corrected video frames.
  
\subsection{Flow Estimation Network}
The purpose of fisheye video correction is to find an intra-frame optical flow for all frame corrections. To achieve this, we first construct a flow estimation network (FSN) to predict the intra-frame optical flow $W$ for each fisheye frame $F$. The FSN is a pyramid structure that uses 6 convolutional layers to extract features and reconstruct optical flows. The input of the network is 5 consecutive fisheye images $\left \{ F_{t-4},F_{t-3},F_{t-2},F_{t-1},F_{t} \right \}$ with size of $256\times256$. The shared-weight FSN makes independent predictions for each frame. The predicted five intra-frame optical flows $\left \{ W_{t-4},W_{t-3},W_{t-2},W_{t-1},W_{t} \right \}$ can be used to describe the deformation of each fisheye frame. Besides, the FSN also outputs the feature map $X$ with a size of $(64,64,32)$ on the encoder. Because independent prediction in the FSN disrupts the temporal correlation, we need to establish the temporal correlation for $X$ in the subsequent flow enhancement network (FEN).
  
\subsection{Flow Enhancement Network}
The predicted intra-frame optical flow varies according to the content of the frame. Then the inconsistency correction on the fisheye frames leads to the video jitter. Therefore, we first design a progressive temporal weighting scheme to perform weighting on $\left \{ W_{t-4},W_{t-3},W_{t-2},W_{t-1},W_{t} \right \}$. The generated unified intra-frame optical flow $W^{'}$ is the input of the flow enhancement network (FEN). Since each $W$ is predicted independently, we introduce a temporal deformation aggregator (TDA) to establish the temporal correlation for the features $X$. In addition, we observe that the difference between the inter-frame optical flow of the fisheye video $M_{t-1 \to t }^{f}$ and preliminary corrected video $M_{t-1 \to t }^{n}$ reflects the local deformation of the frame, as derivation in Equation \ref{deta}. Therefore, we calculate the attention $A$ from $M_{t-1 \to t }^{f}$ and $M_{t-1 \to t }^{n}$ to enhance the local accuracy of $W^{'}$. Our FEN has a 6-layer encoder-decoder structure with a kernel size of $3\times3$ and a stride of 1. Finally, we take $W^{'}$, temporal aggregation information ${X}'$, and spatial deformation feature $A$ as input, and output an improved intra-frame optical flow $W^{\ast }$.

\subsubsection{Temporal Weighting Scheme}
The network outputs different intra-frame optical flows according to different video frames. We need to mitigate the apparent variation between different outputs. Therefore, we design a temporal weighting scheme (TWS). Assuming that we have obtained the intra-frame optical flow of the first fisheye frame $W_{1}$, we treat it as the initial integrated intra-frame optical flow $W^{'}$. When the second flow $W_{2}$ arrives, we need to update $W^{'}$ according to $W_{1}$ and $W_{2}$. We use a simple linear weighting method, which can be expressed by
  \begin{equation}
    W^{'}=aW_{1}+(1-a)W_{2}, \;\; 0<a<1
  \end{equation}
  
To alleviate the jitter effect, it is necessary to make $W^{'}$ tend to $W_{1}$, which implies $a>0.5$. The weight of the subsequent flow should be smaller than the previous flow so that the integrated flow does not change greatly due to the new input. We use a progressive temporal weighting scheme, which can be recorded as
\begin{equation}
W^{'}=a_{1}W_{1}+a_{2}W_{2}+\cdots +a_{n-1}W_{n-1}+a_{n}W_{n}
\end{equation}
where the weight set ${a_{1},a_{2},\cdots,a_{n-1},a_{n}}$ is an arithmetic progression and their sum is 1. In our experiment, $n=5$. We empirically set $a_{1}=0.3$. Therefore, for our intra-frame optical flows $\left \{W_{t-4},W_{t-3},W_{t-2},W_{t-1},W_{t} \right \}$, their corresponding weights are $\left \{ 0.3, 0.25, 0.2, 0.15, 0.1 \right \}$.

\subsubsection{Temporal Deformation Aggregator}
The distortion of the fisheye video increases with the radius. The distorted moving object may be able to find its distorted-free appearance at other times. However, this temporal correlation is broken in the flow estimation network due to independent flow prediction. Therefore, we introduce a temporal deformation aggregator (TDA). With the help of 3D convolution, it performs reconstruction in the temporal dimension for output features of the flow estimation network and obtains a new feature ${X}'$ with temporal correlation. Finally, we concatenate ${X}'$ and the weighted intra-frame optical flow $W^{'}$ on the channel and feed them to the flow enhancement network. ${X}'$ can help the flow enhancement network to improve the global accuracy of the $W^{'}$.

\subsubsection{Spatial Deformation Perception Module}
According to Equation \ref{deta}, the difference between the inter-frame optical flow of the fisheye video $M_{t-1 \to t }^{f}$ and the preliminary corrected video $M_{t-1 \to t }^{n}$ reflects the local deformation. It indicates that we can further enhance the local accuracy of $W_{f}$ through $M_{t-1 \to t }^{f}$ and $M_{t-1 \to t }^{n}$. However, to obtain the inter-frame optical flow of the distorted-free video, it is necessary to perform a preliminary correction on the fisheye video. Therefore, in the spatial deformation perception module (SDM), we first use the intra-frame optical flows $\left \{W_{t-4},W_{t-3},W_{t-2},W_{t-1},W_{t} \right \}$ to correct fisheye frames $\left \{F_{t-4},F_{t-3},F_{t-2},F_{t-1},F_{t} \right \}$ and obtain the rectified frames $\left \{I_{t-4},I_{t-3},I_{t-2},I_{t-1},I_{t} \right \}$. Subsequently, we leverage a lightweight network Raft \cite{teed2020raft} to predict the optical flow for fisheye video $M_{t-n \to t }^{f}$ and the distorted-free video $M_{t-n \to t }^{n}$, respectively. Both $M_{t-n \to t }^{f}$ and $M_{t-n \to t }^{n}$ have a size with $(n-1, 2, w, h)$, where $n$, $w$, $h$ represent the number of frames in a timestamp, the length and width of the video. We use a window with a length of 8 to divide the $M_{t-n \to t }^{f}$ and $M_{t-n \to t }^{n}$ into patches. The patches have a size of $(n-1*win, 2, 8, 8)$, where $win=(w/8)*(h/8)$. As the benefit of the fully connected layer, we map the patches of $M_{t-n \to t }^{f}$ to the query and value, and the patches of $M_{t-n \to t }^{n}$ to the key, to calculate the attention $A$. In the same way, we concatenate $A$ to the flow enhancement network. With the assistance of attention $A$, the network can dynamically perceive the deformation and further enhance the local accuracy of the optical flow.

\subsection{Training strategy}

In the flow estimation network, we use $W_{t}$ to correct $F_{t}$, and obtain a coarse corrected video $Y_{c}$. Since we need to estimate the inter-frame optical flow for the corrected video, the supervision of $Y_{c}$ is necessary. In the flow enhancement network, the outputs $W_{f}$ is leveraged to correct $F_{t}$, resulting in a corrected video $Y_{r}$. We first design a reconstruction loss to minimize the L1 distance between $Y_{c}$, $Y_{r}$ and the ground-truth sequence $\widehat{Y}$. It can be expressed as
\begin{equation}
\mathcal{L}_{R}=\left \| Y_{c}-\widehat{Y} \right \|_{1} + \left \| Y_{r}-\widehat{Y} \right \|_{1}
\end{equation}
  
Our network implements image-to-image transformation using Encoder-Decoder architecture. We thereby use the adversarial loss to align the domains and supervise the final output $Y_{r}$, which can be represented as
\begin{equation}
\begin{aligned}
\mathcal{L}_{A}=\underset{G_{c}}{min}\underset{D}{max} & \left ( {E\left [ logD\left ( \widehat{Y} \right ) \right ] + } \right. \\
 & \left. {E\left [ log\left ( 1-D\left ( G_{c}\left ( Y_{r} \right ) \right ) \right ) \right ]}\right )
\end{aligned}
\end{equation}

In addition, resampling the image with optical flow causes blur, so we enhance the details with style loss $\mathcal{L}_{S}$ and feature loss $\mathcal{L}_{F}$ \cite{StyleLoss} as follow
\begin{equation}
    \mathcal{L}_{S}=\left \| G(\phi (Y_{r}))- G(\phi (\widehat{Y}))\right \|_{F}^{2}
\end{equation}

\begin{equation}
    \mathcal{L}_{F}=\frac{1}{CHW}\left \| \phi \left ( Y_{r}  \right)-\phi\left ( \widehat{Y} \right )  \right \|_{2}^{2}
\end{equation}
Where $\phi(\cdot )$ is the feature of each layer in the VGG16 network, and $G(\cdot )$ is the feature multiplied by its transpose.

Finally, we combine the abovementioned losses to train our network. The overall loss is:
\begin{equation}
  \mathcal{L}={\lambda}_{r}\mathcal{L}_{R} + {\lambda}_{a}\mathcal{L}_{A}  + {\lambda}_{s}\mathcal{L}_{S} + {\lambda}_{f}\mathcal{L}_{F}
\end{equation}
where $\left \{{\lambda}_{r},{\lambda}_{a},{\lambda}_{s},{\lambda}_{f}\right \}$ are the hyperparameter. We empirically set $\left \{20, 0.05, 250, 0.5 \right \}$. 

\begin{figure*}[!t]
  \centering
  \includegraphics[scale=0.165]{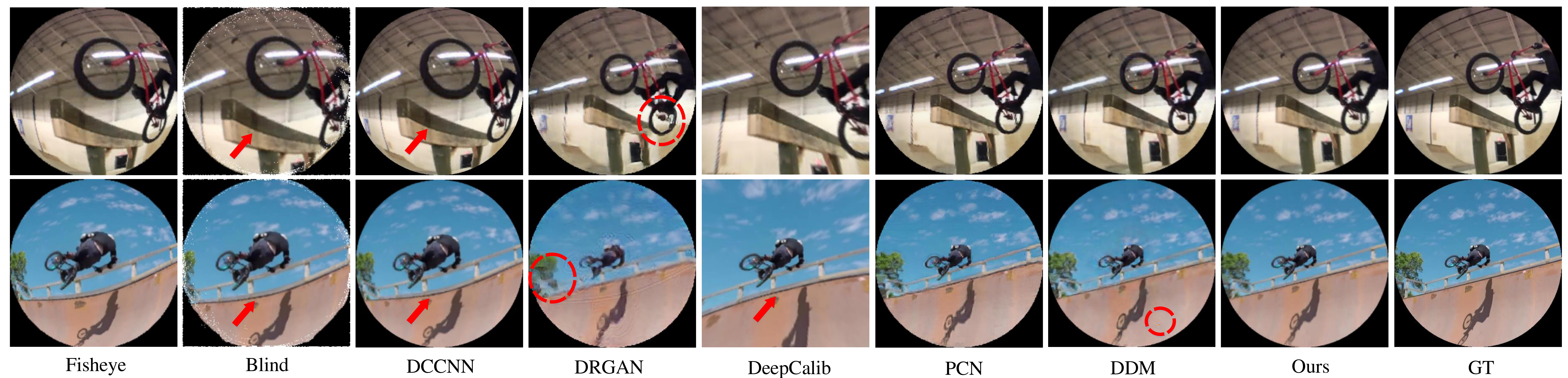}
  \vspace{-0.2cm}
  \caption{
  \label{result}
  Visualization of qualitative comparison results. The first row demonstrates the correction results on Youtube-VOS video sequences, and the rest are the results on DAVIS video sequences. We marked defect structure and content with red arrows and circles, respectively (zoom in to see the details).} 
  \vspace{-0.3cm}
\end{figure*}

\begin{figure*}[!t]
  \centering
  \includegraphics[scale=0.17]{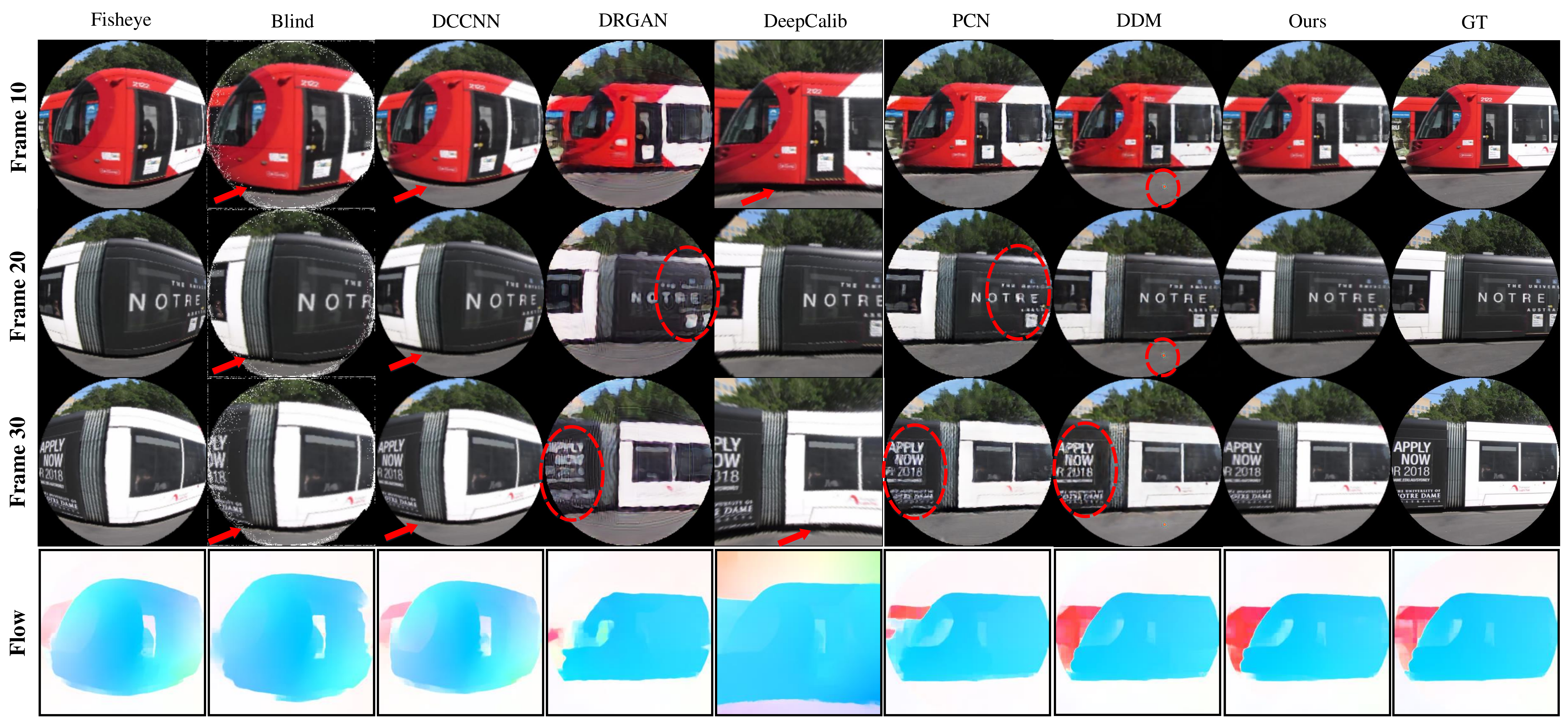}
  \vspace{-0.2cm}
  \caption{
  \label{stability_result}
  The correction result on the fisheye video. Our correction result maintains stability and local quality. Red arrows and circles highlight the distortion and artifact, respectively. Besides, we visualize part of the optical flows (bottom). }
\vspace{-0.4cm}
\end{figure*}

\section{Experiment}
\subsection{Experiment Setup}
Since fisheye video correction has rarely been explored, the generation of our dataset is referred to previous fisheye image synthesis methods \cite{DDM}\cite{PCN}. We first use DAVIS \cite{Davis} to synthesize our fisheye videos. DAVIS includes 90 videos for training and 30 for testing. Additionally, we apply Youtube-VOS \cite{Youtube} which contains 3400 training videos and 500 testing videos. We use random parameters to generate 3.5K training timestamps and 0.1K testing timestamps with the video of DAVIS, and generate 50K training timestamps and 3K testing timestamps with the video of Youtube-VOS. In the experiment, we resize the input timestamp to $256\times256$ and set the batch size to 4. We leverage the Adam optimizer with a learning rate of $1\times 10^{-4}$ to train the network on an NVIDIA GeForce RTX 3090 for 30 epochs.

\begin{table*}[htbp]
  \centering
  \caption{Performance comparison with different methods.}
  \vspace{-0.2cm}
  \setlength{\tabcolsep}{1.3mm}{
    \begin{tabular}{|c||c||c|c|c|c|c|c|c|c|}
    \hline
    \multicolumn{1}{|c||}{Methods} & Dataset & \multicolumn{1}{c|}{PSNR} & \multicolumn{1}{c|}{SSIM} & \multicolumn{1}{c|}{FID} & \multicolumn{1}{c|}{CW-SSIM} & \multicolumn{1}{l|}{Cropping} & \multicolumn{1}{l|}{Distortion} & \multicolumn{1}{l|}{Stability} & \multicolumn{1}{l|}{EPE}\\
    \hline
    \multirow{2}[2]{*}{Blind \cite{Blind}} & DAVIS & 11.33      &  0.3141     &  173.5     & 0.5943 & 0.807      & 0.575      & 0.736 & 3.97\\
          & Youtube-VOS &   11.65    &  0.3164     &   190.8    &  0.5934 &  0.754     &  0.529     & 0.683  &1.23\\
    \hline
    \hline
    \multirow{2}[2]{*}{DCCNN \cite{Rong2016Radial}} & DAVIS &  14.63     &  0.4231     &   119.4    & 0.7222 &  0.813     &  0.398     & 0.733 &3.01\\
          & Youtube-VOS &  14.99     &  0.4252     &  125.7     & 0.7113 &  0.828     &   0.421    & 0.781  &1.01\\
    \hline
    \hline
    \multirow{2}[2]{*}{DRGAN \cite{Liao2019}} & DAVIS &  17.09     &   0.4714    &   194.5    & 0.8269 &   0.822    &  0.673     & 0.726 & 2.01\\
          & Youtube-VOS &   15.67    &   0.4058    &   212.9    & 0.7387 &   0.780    &  0.619     & 0.676  &0.88\\
    \hline
    \hline
    \multirow{2}[2]{*}{DeepCalib \cite{DeepCalib}} & DAVIS &  15.96     &  0.5198     &   73.7    & 0.7841 &  0.911     &   0.677    & 0.767 &7.46 \\
          & Youtube-VOS &   14.72    &   0.4781    &  123.6     & 0.6393 & 0.836      &  0.558     & 0.679  &5.51\\
    \hline
    \hline
    \multirow{2}[2]{*}{PCN \cite{PCN}} & DAVIS &  22.95     &  0.7980     &  86.1     & 0.9537  &   1.000    &  0.916     & 0.869 &1.36 \\
          & Youtube-VOS &  24.98     &  0.8182     &  46.2     &  0.9512 &  1.000     &   0.965    & 0.892  &0.35\\
    \hline
    \hline
    \multirow{2}[2]{*}{DDM \cite{DDM}} & DAVIS &   24.90    &  0.8317    &  63.2     & 0.9614  &  0.988     &  0.957     & 0.877 & 1.58\\
          & Youtube-VOS &  26.42    &  0.7948     &   47.2    & 0.9446 &  0.990     &  0.959     & 0.865  &0.42\\
    \hline
    \hline
    \multirow{2}[2]{*}{Ours} & DAVIS &  \textbf{25.92}     &   \textbf{0.8731}    &  \textbf{45.4}     & \textbf{0.9727} &  \textbf{1.000}     &  \textbf{0.966}     & \textbf{0.888} & \textbf{0.84}\\
          & Youtube-VOS &   \textbf{28.18}    &  \textbf{0.8889}     &   \textbf{37.2}    & \textbf{0.9701}  &  \textbf{1.000}     &  \textbf{0.974}     & \textbf{0.895}  &\textbf{0.24}\\
    \hline
    \end{tabular}}%
  \label{tab_result}%
  \vspace{-0.2cm}
\end{table*}%

\begin{table*}[htbp]
  \centering
  \caption{The structure performance of different key component.}
  \vspace{-0.2cm}
    \begin{tabular}{|l|cccc|ccc|}
    \hline
          & \multicolumn{1}{c}{PSNR} & \multicolumn{1}{c}{SSIM} & \multicolumn{1}{c}{FID} & \multicolumn{1}{c|}{CW-SSIM} & \multicolumn{1}{c}{Cropping} & \multicolumn{1}{c}{Distortion} & \multicolumn{1}{c|}{Stability} \\
    \hline
    $\textbf{w}\; Flow \; Estimation \; Network$   &  21.23     &   0.7334    &   91.5    &   0.9001    & 0.990 & 0.903 & 0.865\\
    $\textbf{w}\; Flow \; Estimation + Enhancement $ & \textbf{28.18}      &  \textbf{0.8889}     &  \textbf{37.2}     &    \textbf{0.9701}   & \textbf{1.000} & \textbf{0.974} & \textbf{0.895} \\
    \hline
    $\textbf{w/o}\; Temporal \; Weighting \; Scheme$ &  26.98     &  0.8770     &  45.9     &    0.9668  & 1.000 & 0.974 & 0.884\\
    $\textbf{w/o}\; Temporal \; Deformation \; Aggregator$ &  27.54     &   0.8826    &  38.1     &   0.9675    & 1.000 & 0.973 & 0.892 \\
    $\textbf{w/o}\; Spatial \; Deformation \; Perception$ &  27.03     &  0.8771     &   46.2     &  0.9663    &  1.000 & 0.972 & 0.891 \\ 
    \hline
    \end{tabular}%
  \label{tab_ablation}%
  \vspace{-0.4cm}
\end{table*}%

\subsection{Qualitative and Quantitative Comparison}
\begin{figure}[!t]
  \centering
  \includegraphics[scale=0.175]{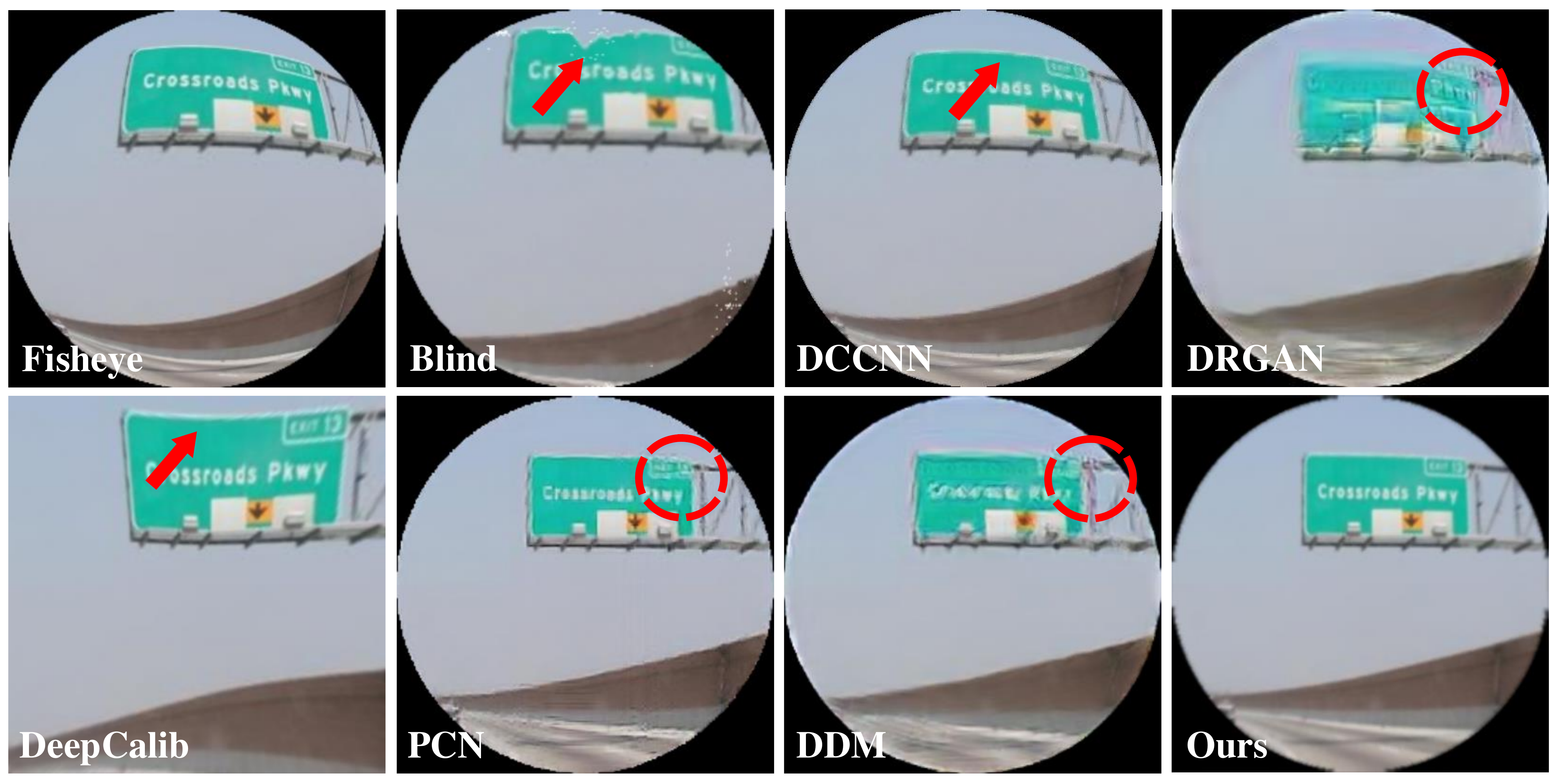}
  \caption{
  \label{real_result}
  Qualitative comparison on real fisheye videos. }
  \vspace{-0.8cm}
  \end{figure}

\begin{figure}[!t]
  \centering
  \includegraphics[scale=0.175]{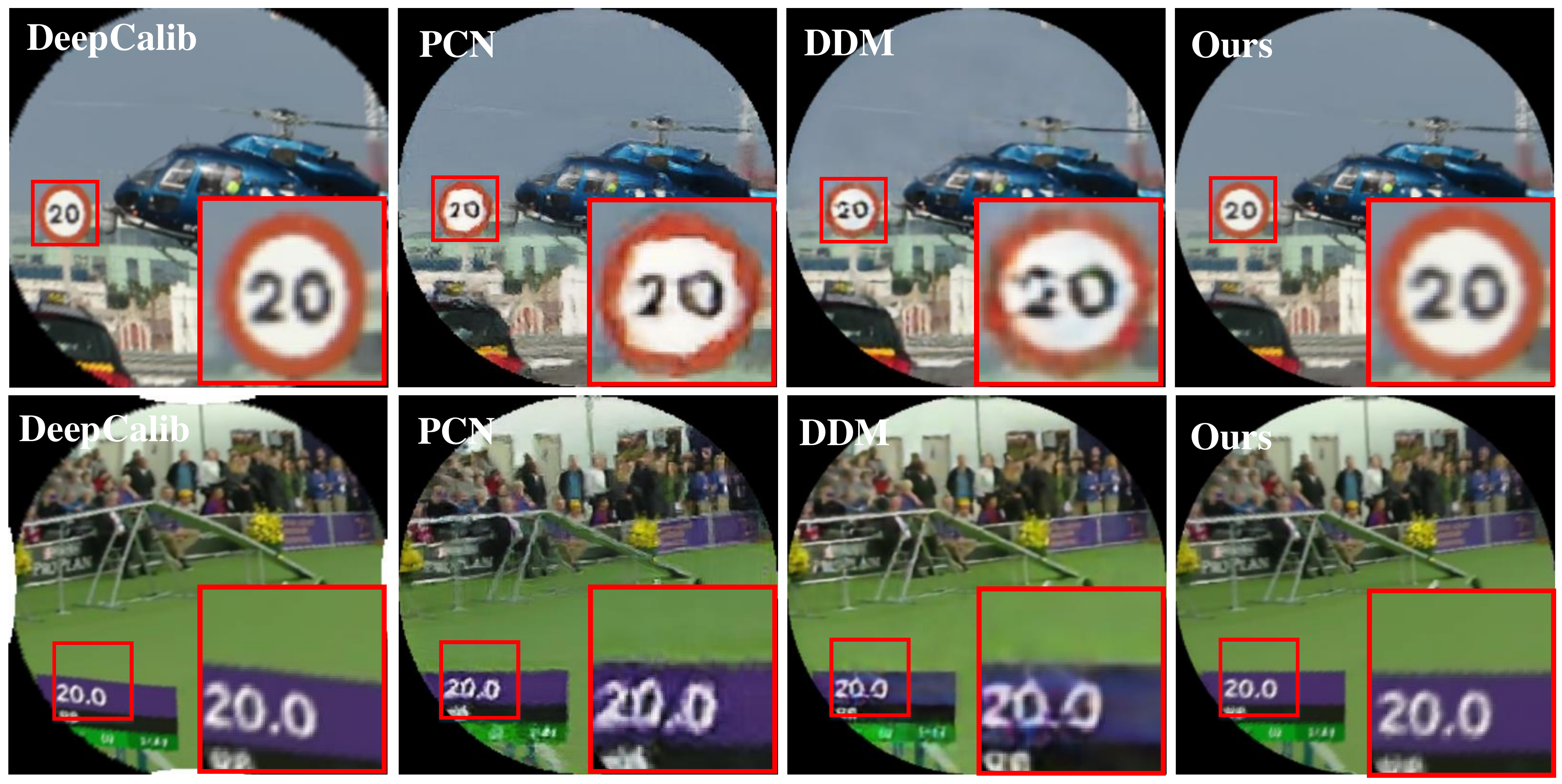}
  \caption{
  \label{zoom_in}
  We zoom in the corrected results to demonstrate that our results have better texture details. }
\vspace{-0.3cm}
\end{figure}

We perform qualitative and quantitative comparisons with state-of-the-art methods Blind \cite{Blind}, DCCNN \cite{Rong2016Radial}, DRGAN \cite{Liao2019}, DeepCalib \cite{DeepCalib}, DDM \cite{DDM}, PCN \cite{PCN}. From Figure \ref{result} and Figure \ref{real_result}, DCCNN and DRGAN produce unsatisfactory results, because their simple structure cannot thoroughly learn the transformation from source domain to the target domain. Blind and DeepCalib improves the structure, but they loss the correction boundary. DDM and PCN make full-frame corrections for each pixel preserved, but their discrete edge points and unsupervised flow lack explicit guidance. Therefore, they produce artifacts as shown in Figure \ref{zoom_in}. In contrast, our method quickly construct the guidance of flow by supervising two transformed video frames. Subsequently, the spatiotemporal deformation perception further enhance the global and local accuracy of flow. Our method thereby achieves better visual performance with accurate structure and realistic texture. 

In addition, we also calculate PSNR (Peak Signal to Noise Ratio), SSIM (Structural Similarity), FID (Frechet Inception Distance) \cite{FID}, and CW-SSIM (Complex Wavelet Structural Similarity) \cite{CW_SSIM} to objectively evaluate the corrected videos. The comparison of quantitative results in Table \ref{tab_result} intuitively highlights the superior performance of our method.

\subsection{Stability Analysis}
\label{st_analysis}
To verify the stability of our approach, we also test on fisheye videos. Since the most accurate prediction of intra-frame optical flow is difficult to obtain, we therefore use Blind, DCCNN, DRGAN, DeepCalib, DDM, and PCN to independently predict and correct each frame. The comparison method and our method have one and five input frames, respectively. For a fair comparison, our method also makes independent predictions for each frame. In a video, we generate multiple timestamps with overlapping by assuming the first timestamp is $\left \{F_{t-4},F_{t-3},F_{t-2},F_{t-1},F_{t} \right \}$ and the next timestamp is $\left \{F_{t-3},F_{t-2},F_{t-1},F_{t},F_{t+1} \right \}$. Subsequently, we take the last frame of each corrected timestamp as the result of an independent prediction. Figure \ref{stability_result} shows the correction results of some frames in the same fisheye video. It can be seen that DCCNN, DRGAN, Blind, DeepCalib, DDM, and PCN have large differences in the correction of different frames, while our method achieved better stability with the help of a temporal weighting scheme. In order to compare the stability more intuitively, we use common evaluation metrics (cropping, distortion, and stability) \cite{st_metric} in video stabilization to measure the stabilization performance of the corrected video. In Table \ref{tab_ablation}, our method is higher than other methods in cropping, distortion, and stability, which further indicates the effectiveness of our method.
  
\subsection{Testing on Downstream Tasks}
Testing on downstream tasks can verify the quality of the corrected video. Therefore, we use Raft \cite{teed2020raft} to estimate the optical flow from corrected videos (in Figure \ref{stability_result}), and calculate their EPE (end-point-error) (in Table \ref{tab_result}). Our method has the lowest EPE, which reflects that our corrected video has the most promising quality.
  
\subsection{Ablation Study}
Our network consists of multiple modules. We design several experiments to verify the effectiveness of each key component. The experimental results are shown in Table \ref{tab_ablation}. When the network contains only the flow estimation network, the experimental results are unsatisfactory. A simple network model does not predict optical flow accurately. When the flow enhancement network is introduced, the network obtains a more accurate optical flow by further perceiving the temporal information and local deformation, thereby improving the performance. Compare the second and third rows, removing the temporal weighting scheme reduces the stability, which explicitly indicates the effectiveness of the scheme. From the second and last two rows, when the temporal deformation aggregator and the spatial deformation perception module are unavailable, the change of the stability is not obvious but the quality performance is greatly degraded. It proves that both temporal information and local deformation benefit optical flow estimation.
  
\section{Conclusion}
In this paper, we design a novel network to correct the fisheye video for the first time. Although the existing methods provide satisfactory results on the image, the problem of jitter will be incurred on the video correction. Thus, we first propose a temporal weighting scheme to alleviate the jitter caused by the new input frames. Then, we introduce a temporal deformation aggregator to establish the temporal correlation of features, thereby improving the global accuracy of the intra-frame optical flow. Besides, our spatial deformation perception module can help the network perceive the deformation of the object and improve the local accuracy of intra-frame optical flow. Compared to the state-of-the-art methods, our proposed method achieves the best performance in terms of correction quality and stability. In the future work, we will improve the method to meet the real-time requirements.

\section{Acknowledgments}
This work was supported by  the National Natural Science Foundation of China (Nos. 62172032,  62120106009).

\bibliography{ref}

\end{document}